%% file: main.tex
\documentclass[11pt,a4paper]{article}
\usepackage[dvipsnames]{xcolor}
\usepackage{times,latexsym}
\usepackage{url,linguex}
\usepackage[T1]{fontenc}

\usepackage[acceptedWithA]{tacl2021v1}
\usepackage{soul}
\usepackage{amsmath}

\usepackage{xspace,mfirstuc,tabulary,todonotes}

\usepackage{color}
\usepackage{tcolorbox}
\usepackage{adjustbox}
\usepackage{booktabs}
\usepackage{enumitem}
\usepackage{caption}

\newcounter{taxexample}
\renewcommand{\thetaxexample}{Ex.\arabic{taxexample}}
\newcommand{\taxex}[1]{\leavevmode\refstepcounter{taxexample}\label{#1}\thetaxexample}

\newcounter{responsecounter}[taxexample]
\renewcommand{\theresponsecounter}{\arabic{taxexample}.\arabic{responsecounter}}
\newcommand{\response}[1]{\leavevmode
  \refstepcounter{responsecounter}\label{#1}\textcolor{Plum}{Resp.\theresponsecounter:}}

\newcommand{\nliex}[3]{
  \textcolor{Emerald!90!black}{\textbf{P:} #1}
  \textcolor{Emerald!90!black}{\textbf{H:} #2} \hfill\livenli [E,N,C]: #3
}
\newcommand{\ContextStatement}[2]{
  \textcolor{Emerald}{\textbf{Context:} #1}
  \textcolor{Emerald}{\textbf{Statement:} #2}
}
\newcommand{\phpair}[2]{
  \textcolor{Emerald!90!black}{\textbf{P:} #1}
  \textcolor{Emerald!90!black}{\textbf{H:} #2}
}

\newcommand{\QUD}[1]{\textcolor{gray}{QUD: #1}}

\definecolor{entail_blue}{rgb}{0.12, 0.46, 0.71}
\definecolor{neutral_orange}{rgb}{1.0, 0.49, 0.05}
\definecolor{contradiction_green}{rgb}{0.17, 0.62, 0.17}

\newcommand{\Entailment}{\textcolor{Plum}{Entailment} \xspace}
\newcommand{\Neutral}{\textcolor{Plum}{Neutral} \xspace}
\newcommand{\Contradiction}{\textcolor{Plum}{Contradiction} \xspace}

\newif\iftaclinstructions
\taclinstructionsfalse %
\iftaclinstructions
\renewcommand{\confidential}{}
\renewcommand{\anonsubtext}{(No author info supplied here, for consistency with
TACL-submission anonymization requirements)}
\newcommand{\instr}
\fi

\iftaclpubformat %

\else

\fi

\newcommand{\livenli}{\textsc{LiveNLI}\xspace}

\title{Understanding and Predicting Human Label Variation\\ in Natural Language Inference through Explanations}

\author{

  Nan-Jiang Jiang$^1$ \quad Chenhao Tan$^2$ \quad   Marie-Catherine de Marneffe$^{13}$ \\
  $^1$ Department of Linguistics, The Ohio State University, USA\\
  $^2$ University of Chicago, USA\\
  $^3$ FNRS, UCLouvain, Belgium\\
  \texttt{jiang.1879@osu.edu} \quad
  \texttt{chenhao@uchicago.edu} \quad
  \texttt{demarneffe.1@osu.edu} }

\date{}

\begin{document}
\setlength{\Exlabelsep}{.5pt}
\setlength{\Exlabelwidth}{2em}
\setlength{\Extopsep}{0.4\baselineskip}
\maketitle
\begin{abstract}
Human label variation \cite{plank-2022-problem}, or annotation disagreement, exists in
many natural language processing (NLP) tasks.
To be robust and trusted, NLP models need to identify such variation and be able to explain it. To this end, we created the first ecologically valid explanation dataset with diverse reasoning, \livenli.
\livenli contains annotators' highlights and
free-text explanations for the label(s) of their choice for 122 English Natural Language Inference
items, each with at least 10 annotations. %
We used its explanations for chain-of-thought prompting, and found
there is still room for improvement in GPT-3's ability to predict label
distribution with in-context learning.
\end{abstract}

\section{Introduction}
\label{sec:intro}
Until recently, practices for operationalizing annotations in natural language processing (NLP) datasets assumed one single label per item. However, human label variation \citep{plank-2022-problem} has been found in a wide range of NLP tasks, including
 part-of-speech tagging, coreference resolution, and natural language inference (NLI) \citep[i.a.]{plank2014learning,poesio-etal-2018-anaphora,pavlick-kwiatkowski-2019-inherent,nie-etal-2020-learn,uma2021learning,qp2}. \citet{aroyo_truth_2015}, among others, argued that such variation in annotations should be considered signal, not noise.
Specifically, the NLI task --- identifying whether the hypothesis is true (Entailment), false (Contradiction), or neither (Neutral) given a premise ---
 has embraced label variation and set out to predict it \cite{zhang-etal-2021-learning,zhou-etal-2022-distributed}.
 However, the question of where label variation in NLI stems from remains open.

\input{table_example}

As an initial effort to tackle this question,
\citet{qp2} %
introduced a taxonomy of linguistic phenomena that can lead to different interpretations of English NLI items,
and thus potentially to different NLI labels.
For instance, in \ref{ex:pundit} (Table~\ref{tab:examples}), does one infer \textit{a large following} in the hypothesis from \textit{most} in the premise?
This lexical indeterminacy was hypothesized by \citet{qp2} to have contributed to the variation that is reflected in the label distribution.

\citet{qp2}'s taxonomy was built post-hoc, and hence detached from the annotators who provided the NLI labels.
It is thus important to understand the reasons from the perspective of annotators and validate their taxonomy.
Furthermore, given that human label variation is widespread and systematic, NLP models that can predict and explain label variation would be useful for improving annotation and downstream utility.

We introduce \livenli, Label Variation and Explanation in NLI.
\livenli contains 122 re-annotated MNLI
\citep{williams-etal-2018-broad} items, each with at least 10 annotations, including highlights and free-text explanations for the labels chosen by the annotators. Compared to previous explanation datasets, \livenli offers more diverse and ecologically valid explanations as each annotator explained the label they chose.

\livenli provides direct evidence for how label variation among annotators arises,
allowing us to compare annotators' explanations against \citet{qp2}'s taxonomy.
The free-text explanations in \livenli showed that the reasons for label variation are largely similar to
the taxonomy, but other reasons also emerge.
Further, we found \textit{within-label variation}: even when annotators chose the same label, they may have different reasons to do so. %
Our highlights analysis also confirmed \citet{tan-2022-diversity}'s observation that highlights by themselves do not provide effective explanations.

Finally, we explore the promise of using NLP models to predict human label
variation, and \livenli allows us to introduce explanations as additional supervision.
Inspired by chain-of-thought prompting~\citep{wei_chain_2022}, we show that incorporating explanations in the prompt slightly improves the ability of GPT-3 \citep{gpt3} to predict NLI label distributions over prompting without explanations. %
Yet, there is still room for improvement in both predicting distributions and generating explanations.\footnote{The data and code will be publicly available on GitHub.}

\section{Related Work}
\subsection{Label Variation in NLI}
Previous work has found systematic label variation in NLI datasets
\citep{de-marneffe-etal-2012-happen,pavlick-kwiatkowski-2019-inherent,nie-etal-2020-learn,uma2021learning}.
\citet{qp2} proposed a taxonomy of 10 linguistic categories that can lead to label variation.
\ref{ex:pundit} illustrated the ``Lexical'' category.
We refer the reader to \citet{qp2} for the full definition of all 10 taxonomy categories.

In this work, we validate and improve the taxonomy based on
the explanations annotators provide when choosing their label(s).
The taxonomy categories do not specify what the different readings are for the different labels, e.g.,\ in the ``Lexical'' category, which words carry uncertain meaning.
Another frequent taxonomy category in \citet{qp2} is
``Probabilistic Enrichment'':
some annotators make different probabilistic inferences from the premise that
are likely but not definitely true. In \ref{ex:summer} (Table~\ref{tab:examples}), Responses~\ref{ex:summer_response2} and
\ref{ex:summer_response_N} make different inferences about whether
the speaker likes summer given that they dislike winter: \ref{ex:summer_response2} considers that the speaker likes it because of the live foliage, while \ref{ex:summer_response_N}
hypothesizes that the speaker might dislike summer because of the heat.
The category ``Probabilistic Enrichment'' by itself
does not tell us what the probabilistically enriched content is.
The explanations, on the other hand, do convey this information, and would be more useful for non-experts to understand label variation.

\paragraph{Predicting NLI label distribution}
\citet{pavlick-kwiatkowski-2019-inherent} argued that models should
preserve information about label variation by predicting label distributions to propagate such information to downstream tasks.
\citet{zhang-etal-2021-learning} and \citet{zhou-etal-2022-distributed}
introduced BERT-based fine-tuned models for predicting NLI label
distribution, using the ChaosNLI dataset \cite{nie-etal-2020-learn} which
contains 100 annotations/item.
Here, we explore predicting label distributions with in-context learning
with large language models, namely \ GPT-3 \cite{gpt3}, using
explanations as additional supervision.

\subsection{Explanation Datasets}
With the rise of interest in interpretability in NLP, many datasets where
labels are given explanations have been introduced (see \citet{wiegreffe2021teach} for a survey).
The dataset most similar to \livenli is e-SNLI
\citep{NEURIPS2018_4c7a167b}: it also targets the NLI task and contains free-text explanations and
highlights for all items in SNLI \cite{bowman-etal-2015-large}.
e-SNLI has been extensively used for building models that produce explanations
\citep[i.a.]{kumar-talukdar-2020-nile,narang_wt5_2020,
  zhao-lirex-2021,wiegreffe-etal-2022-reframing,yordanov_few-shot_2021} and
studying whether models can learn from explanations to perform the classification task
\citep[i.a.]{wiegreffe-etal-2021-measuring,hase-bansal-2022-models,ye_unreliability_2022}.
Our work extends e-SNLI on several dimensions.

\paragraph{Ecological validity}
An explanation is ecologically valid if the same annotator provides both the label to
explain and the explanation.
Many explanation datasets are not ecologically valid: instead of soliciting both the label and the explanation, only the explanations for
``why this item has this label'' were collected, and the label to explain is the ``ground truth'' label collected in previous work.
However, given that label variation is widespread in many NLP tasks,
only explaining the ground truth is not capturing the
full range of meaning and may not reflect the annotators' decision making process.
This also poses a problem for evaluating explanations, as
\citet{wiegreffe-etal-2022-reframing} found that humans have biases against
explanations when they disagree with the label.
In this work, we ask annotators to label NLI items and explain their chosen label, ensuring the ecological validity of the explanations.

\paragraph{Diversity of the explanations}

We aim to collect explanations that are diverse on multiple aspects. First, as discussed earlier, most existing explanations seek to explain the ``ground
truth'' label without taking into account human label variation.
By focusing on explaining items that are known to exhibit variation,
explanations in \livenli reflect how the same item can have different labels.

Second, most explanation datasets for NLP tasks with free-text explanation contain only one explanation per item.
The few large scale datasets with more than one explanation have up to 5
explanations/item \citep{NEURIPS2018_4c7a167b,sap-etal-2020-social,zhang-etal-2020-winowhy,DBLP:journals/corr/abs-2004-03744}.
For \livenli, we collected 10 explanations per item, often explaining more than one label.
This provides us with rich information about reasons behind each label. As we will see, people can indeed arrive at the same label for different reasons.

Lastly, \livenli is based on MNLI \cite{williams-etal-2018-broad}, whereas e-SNLI is based on SNLI. As pointed out by \citet{williams-etal-2018-broad}, SNLI describes visual scenes only, making the hypotheses short and simple,
and non-visual linguistic phenomena, e.g.\ temporal reasoning and
beliefs, rare.
Therefore, the e-SNLI explanations also tend to be simple and templated \cite{camburu-etal-2020-make,tan-2022-diversity}, and are unlikely to exhibit diverse types of reasoning.
We chose MNLI to ensure that the items involve different types of reasoning, which in turn
contributes to having diverse explanations.

\paragraph{Learning from explanations}
Previous work did not find much success incorporating explanations to improve
classifiers using datasets such as e-SNLI with fine-tuning settings \citep{NEURIPS2018_4c7a167b,wiegreffe-etal-2021-measuring,hase-bansal-2022-models}.
Chain-of-thought prompting, a.k.a.\ Explain-then-Predict \cite{wei_chain_2022,ye_unreliability_2022}, on the other hand, improves large language model's ability to perform some reasoning
tasks when prompting the model with explanations for the answer before generating it.
We explore using \livenli's explanations to improve GPT-3's ability to predict label variation.

\section{Data Collection}
We re-annotated 122 items from the MNLI dev set: 110 are
analyzed in \citet{qp2} -- 50 of them are also in the ChaosNLI-re-annotation
\cite{nie-etal-2020-learn}, with 100 annotations/item.
The other 12 items (4 for each NLI label) have unanimous agreement in the original MNLI 5 annotations. We intended to use those for quality control, to filter annotators who
did not give the same label as the original one and might therefore not be
paying attention. However, as discussed below, those items also exhibit
systematic label variation with legitimate explanations.

\begin{figure*}[t]
  \centering
 \includegraphics[width=\linewidth]{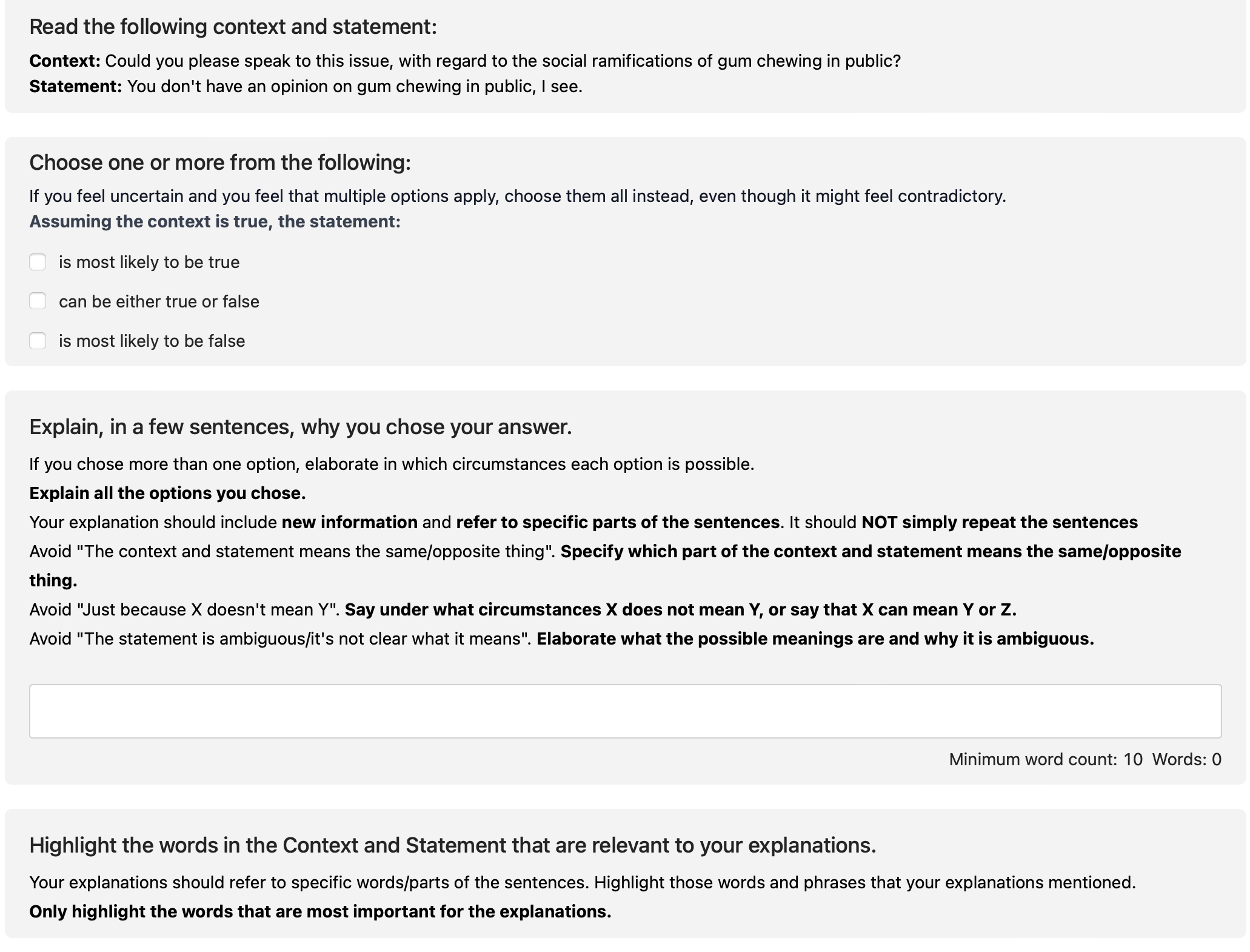}
  \caption{Data collection interface.}
  \label{fig:interface}
\end{figure*}

\paragraph{Procedure}
Figure~\ref{fig:interface} shows a screenshot of the annotation interface on the Surge AI crowdsourcing annotation platform.\footnote{\url{surgehq.ai}} Annotators read the premise (context) and hypothesis (statement), and were asked whether the statement is most likely to
be true/most likely to be false/either true or false (corresponding to
Entailment/Contradiction/Neutral, respectively). They can choose multiple labels if they feel uncertain. They were then asked to provide a free-text explanation for all the label(s) they
chose. They were instructed to give explanations that provide new information and refer to specific parts of the sentences, and avoid to simply repeating the sentences.
These instructions were inspired by previous work that identified features of high and low quality explanations
\citep{camburu-etal-2020-make,wiegreffe-etal-2022-reframing,tan-2022-diversity}.
Annotators were further instructed to highlight words from the premise and
hypothesis that are most important for their explanations. Surge AI guarantees
that their annotators are native English speakers, as well as their reading and
writing abilities. Annotators were paid \$0.6 per item (hourly wage:
\$12 to \$18). Each item received at least 10 annotations.
48 annotators participated in total.
On average, each annotator labeled 29.3 items.
Our data collection was performed with IRB approval.

\section{Label Analysis}
We analyzed the multilabel annotations to quantify the amount of
variation and the extent to which they align with the original MNLI annotations.

\paragraph{Descriptive statistics}
In 83\% of the label responses, only one label was chosen, with 53\% of all
responses being Entailment or Contradiction. The majority of the annotators thus
express clear judgments. Much of the label variation arises from
variation across individuals
as opposed to each individual having uncertain or gradient judgments.

\begin{figure}[t!]
  \centering
  \includegraphics[width=0.9\linewidth]{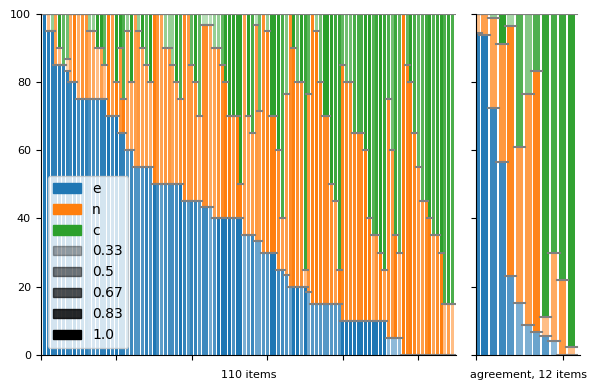}
  \caption{
    Average distributions from the normalized multilabel responses for each
    item.
    For each item and each label, we take the multilabel responses that include
    at least that label and compute the average probabilities for that
    label (bars' transparency).
    Label variation is widespread, even for the items with unanimous agreement in MNLI.
  }
  \label{fig:stacked}
\end{figure}

To aggregate the labeling responses for each item,
we first normalize each multilabel response into a distribution, so that if an
annotator chose multiple labels, each label will be weighted less
than if they only chose a single label.
We then average the individual distributions to obtain the item's label distribution.
In Figure~\ref{fig:stacked}, each set of stacked bars shows the distributions
for one item. The transparency of each bar indicates the mean non-zero probabilities for that label.
Figure~\ref{fig:stacked} shows that there are fine-grained
judgments that we need to predict in the form of label distributions.
On the other hand,
there can be similar distributions (shown by the lengths of the bars)
coming from different compositions of multilabels (shown
by the transparency of the bars).

\paragraph{Agreement items}
We intended to use 12 items, originally with unanimous agreement in MNLI, for
quality control.
For each of these 12 items, the labels with the highest
probability in the aggregated distribution
from our annotations are the same as the
unanimous one in the original MNLI annotations, suggesting that our annotators' judgments are reliable.
However, as the right panel of Figure~\ref{fig:stacked} shows, there is a wide
range of variation for those items, but annotators provided legitimate
explanations for the
labels that differ from the original one (see \ref{ex:wax} in
Table~\ref{tab:examples}).

\paragraph{Takeaways}
We found that there is systematic label variation in the \livenli annotations,
including for items that were unanimously agreed on in the original MNLI annotations.
This suggests that label variation is widespread, and we need more annotations
to capture it.
\citet{baan-etal-2022-stop} pointed out that considering one distribution to govern all judgments from the population is a simplification. Our findings confirm that the distributions collected from a pool of annotators (a sample of the population) may shift with a different pool.\footnote{Comparing with the label distributions in ChaosNLI reinforces this statement. We omit the analysis with ChaosNLI here for space reasons.}

\section{Free-text Explanations Analysis}
\citet{qp2} proposed a taxonomy of reasons for label variation in NLI.
Here we qualify the diversity of the \livenli explanations, and analyze the explanations to see what are annotators' reasons for label
variation -- and in particular if they are similar to what \citet{qp2} hypothesized.

\subsection{Quantitative Analysis}
\begin{figure}[t]
  \centering
  \includegraphics[width=0.9\linewidth]{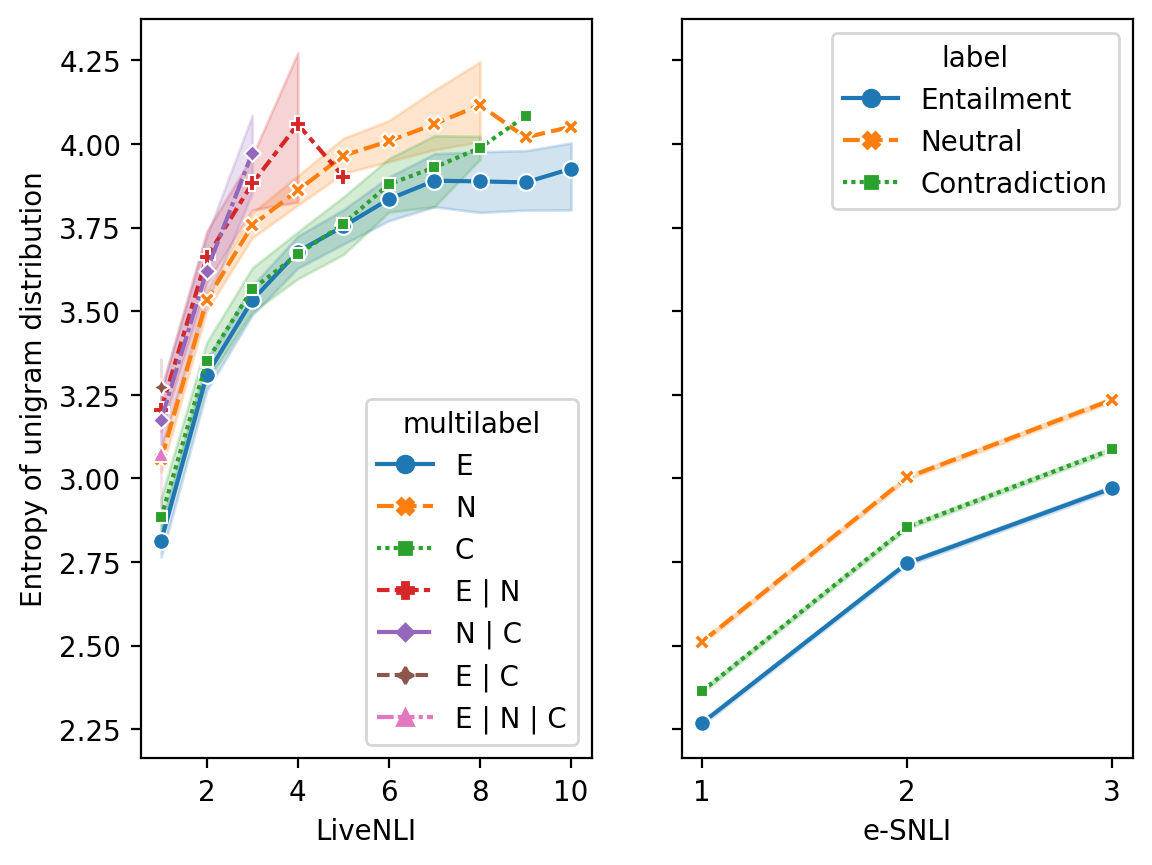}
  \caption{Entropy of distributions of unigrams in the explanations of each item
    with each label (error bars indicating the variability across items.)}
  \label{fig:unigram_entropy}
\end{figure}

For each item $i$ with each (multi)label $l$, Figure~\ref{fig:unigram_entropy} shows the
entropy of unigram distributions calculated using  $i$'s explanations for $l$,
and all the subsets of those explanations (sizes on the
x-axis), with error bars indicating the variability across items $i$.
We also perform the same calculation for the explanations from the e-SNLI dev and test sets.
Both datasets have the same ordering in the entropy of explanations for single
labels: Neutral > Contradiction > Entailment.
This is expected: to explain Entailment, annotators reiterate the premise/hypothesis and tend to use the same words used in the sentences (as in Response \ref{ex:pundit_response_E}), while
to explain Neutral and Contradiction,
annotators need to mention information not already given and are thus likely to use different words (as in Responses~\ref{ex:pundit_both_QUD} and \ref{ex:pundit_response_N}).

Compared with e-SNLI, our explanations have higher unigram distributions entropy, even for the same number of explanations (3/item), showing that our explanations
are more lexically diverse.
They are also longer, with mean token length 28.7 vs.\ 13.3 for e-SNLI.
Focusing on the left panel, we see that entropy increases with the number of explanations: it is not surprising that more explanations increases lexical diversity. But importantly,
the entropy plateaus after 5 explanations for all labels,
which could be the sweet spot for balancing between diversity and annotation budget.

\subsection{Verifying the Taxonomy}
\label{sec:validating_taxonomy}
The explanations in \livenli allow us to investigate the reasons for which annotators chose
different labels.
We investigate how often \citet{qp2}'s categories for label variation appear in the explanations.
For each pair of premise/hypothesis,
one author examined its explanations and assigned the pair one or more categories
of variation exhibited in its explanations.\footnote{For some items, annotators
  unanimously agree on one label and
on their reasons for the label. Those are assigned ``No variation'' in Table~\ref{tab:taxonomy_reannot}.}

Table~\ref{tab:taxonomy_reannot} shows the frequency of the taxonomy re-annotations.
We can see that all the taxonomy categories from \citet{qp2} are used.
Across the board, label variation arises in similar reasons as
what \citet{qp2} hypothesized.
Notably, the two most frequent categories are Probabilistic and Lexical.

\input{table_taxonomy_reannot}

Meanwhile, we found that
many instances of label variation stem from annotators
judging the truth of different at-issue content, which answers different
Questions Under Discussion (QUDs) \citep{roberts_information_2012}.
For example, in \ref{ex:pundit},
Responses \ref{ex:pundit_response_E} and (the first half
of) \ref{ex:pundit_both_QUD} take
the main point of the hypothesis to be Stephanopoulos having a very
large pundit following, but have different judgments on whether this main
point is true.
On the other hand, \ref{ex:pundit_response_N} and \ref{ex:pundit_response_C}
 focus on the reason for which pundits follow Stephanopoulos, and agree
with \ref{ex:pundit_response_E} that he has a large following.
We thus added a ``QUD'' category to the taxonomy (generalizing two of \citet{qp2}'s categories, ``accommodating minimally added content'' and ``high overlap'', which involve annotators having different readings and ignoring certain parts of the items). %
The QUD category occurred in 28 items (out of 122) and is the
third most frequently used category. QUD is thus an important aspect of language understanding that people pick up on. Incorporating QUD
into NLP task designs and modeling \citep{de-kuthy-etal-2018-qud,https://doi.org/10.48550/arxiv.2207.00397} is an interesting avenue of research.

\subsection{Within-Label Variation}
\label{sec:diff_reasons_same_label}
Even though the taxonomy categories were meant to be reasons for variation in labels,
\livenli's explanations show that
there is also \textit{within-label variation}:
annotators can vary in their understanding of the text and have different reasons for the same label.
16 (out of 122) items exhibit such within-label variation, which the taxonomy can capture. We discuss here
two categories in which within-label variation occurs.

\paragraph{Different QUDs}
In \ref{ex:planes} (Table~\ref{tab:examples}), some annotators agree on the Neutral label but their explanations show that they take the main point of the sentence to be different: Response~\ref{ex:planes_encourage} focuses on whether the view being encouraged
is true, while \ref{ex:planes_all} takes the view to be true while questioning whether the planes are all it takes.

\paragraph{Different coreference assumptions}
In \ref{ex:summer}, the explanations indicate that the annotators differ in whether to take the premise and hypothesis to be referring to the same entity/event.
Response~\ref{ex:summer_response1} assumes that the the premise and hypothesis refer to the same season: thus since the premise talks about winter and not summer, the hypothesis \textit{I'm so sick of summer} is false. %
\ref{ex:summer_response2}, on the other hand, does not assume that the premise and the hypothesis refer to the same season, and infers through probabilistic enrichment that the speaker likes summer, making the hypothesis false.
Both with and without the coreference assumption,  annotators label the item as Contradiction.

To summarize, explanations reveal the kinds of variation that labels themselves do not capture.
\livenli will be a useful resource to study variation in human understanding
of texts in general, not limited to variation in labels.

\section{Highlights Analysis}
\input{table_highlight.tex}
Highlights have been criticized in previous work as an inadequate form of explanation
\citep{tan-2022-diversity} because they only provide evidence for the
label without conveying the mechanisms for how the evidence leads to
the label.
Here, we evaluate the quality of the highlights in \livenli, and analyze what information they convey.

\paragraph{Relationship between highlights and labels}

One might hypothesize that
annotators would agree on the highlights when they agree on the label.
To measure agreement on highlights,
for each item,
we computed the overall Krippendorff's $\alpha$ on whether annotators highlight each
word using all the highlights for the item, as well as 7 label-specific
$\alpha$'s using the subset of highlights that agreed on the (multi)label.\footnote{155 responses were not included to calculate the label-specific
  $\alpha$s because
  there is no other responses for the respective item choosing the same multilabel.
}
The mean overall $\alpha$ across all items is 0.22,
while the label-specific $\alpha$'s range from 0.25 to 0.31, suggesting low agreement.
In \ref{ex:wreck} and \ref{ex:tommy} (Table~\ref{tab:highlight}), annotators unanimously agreed on
the label while giving different highlights.
\citet{sullivan-etal-2022-explaining} similarly found low agreement on
highlights for sentiment analysis ($\alpha$ = 0.3), even though the agreement on the sentiment label was high ($\alpha$ close to 1).

On cases where annotators do agree on the highlights,
they can disagree on the label.
In fact, the same highlights can be given for different labels.
In \ref{ex:simile}, three
annotators provided the exact same highlights (only the words
\textit{simile} and \textit{metaphor}), while providing three different labels.
They all consider the words \textit{simile} and \textit{metaphor} to be
important but have different understanding of the relationship between similes and metaphors.
This suggests that, for variation in NLI,
the label depends not only on which words are important, but crucially on
the relationship between the words.

\begin{figure}
  \centering
  \includegraphics[width=0.9\linewidth]{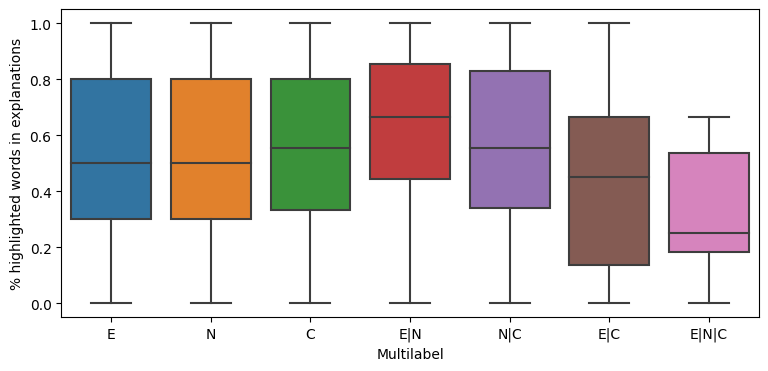}
  \caption{Proportion of highlighted words that are also mentioned
  in the explanation.}
  \label{fig:word_in_exp}
\end{figure}

\paragraph{How grounded in the explanations are the highlights?}
Figure~\ref{fig:word_in_exp} shows the boxplot of highlights and explanation overlap  (percentages of words
highlighted that appear in the corresponding explanation) organized by the (multi)label they explain.
On average, the explanations mention 57\% of the highlighted words.
When annotators chose all the labels (E|N|C), the percentage of
highlighted words is much lower: the free-text explanations mention that
the annotators find the sentences hard to understand and that they cannot make an informed judgment.
This further suggests that the highlights are noisy.
Many highlights include large spans of words covering entire clauses, while only a few words in the clause are discussed in the explanation.
\citet{sullivan-etal-2022-explaining} found that removing drag affordances from
the user interface (forcing annotators to click each word individually)
increases inter-annotator agreement. We used a top-of-the-shelf web interface
provided by Surge AI that allows drag affordances, which
may have contributed to the noise we found.

\paragraph{What leads to high agreement on highlights?}
The mean highlights and explanation overlap for each item
is positively correlated with
the highlight overall $\alpha$ for each item
(Pearson's $r$ = 0.38).
The items with high overlap and high $\alpha$ tend to have
hypotheses with parallel structures to the premise while differing
minimally in certain words: the pair of words that differ are then highlighted and discussed in the explanations.
For example, \textit{leaned-hovered} in \ref{ex:tommy}
and \textit{metaphor-simile} in \ref{ex:simile} are consistently highlighted.

\paragraph{Takeaways}
Our analysis reiterates previous criticism that highlights do not effectively explain labels.
In particular, the same highlight can be used to explain different labels.
For complex tasks such as label variation in NLI,
the free-text explanations are crucial for explaining what the relationships are
between the highlighted words and between the highlights and the labels.

\section{Predict Label Variation by Learning from Explanations}
Given that there is systematic label variation in data,
models need to be able to recognize such variation to obtain human-like understanding.
Explanations have been found to improve large language models' performance on various tasks \cite{wei_chain_2022}.
Therefore, we investigate to what extent \livenli's explanations can be used to help large language models identify label variation.

\paragraph{Setup}
We use the latest variant of GPT-3 \cite{gpt3}: text-davinci-003 with the in-context learning paradigm.
We prompt the model with training items, including the premise/hypothesis, the label distribution in prose, and the explanations for each label,
followed by a test item, consisting of the premise/hypothesis only, without
any label distribution or explanations.
Table~\ref{tab:prompt} shows the prompt for one training item.

\begin{table}
  \centering
  \footnotesize
  \begin{tabulary}{1.0\linewidth}[h]{p{\linewidth}}
    \toprule

  \textcolor{Emerald!90!black}{\textbf{Context:}
Although it is a significant part of the poverty population, Asians
    historically have not been able to participate in the services and programs
    available to the poor, he said. }\\
  \textcolor{Emerald!90!black}{\textbf{Statement:}
Asians are usually not poor. }\\
\textcolor{black}{\fbox{Prediction}}
    Given the context,
    the probability of the statement being \textcolor{Plum}{\bf true is 0.0},
    the probability of the statement being \textcolor{Plum}{\bf false is 0.85},
    the probability of the statement being \textcolor{Plum}{\bf either true or false is 0.15}.\\
\fbox{Explanations}
    \textcolor{Plum}
{\bf It can be false because}
the context notes that Asians are typically a significant part of the poverty population whereas the statement notes that Asians are not usually poor. These two statements seem to contradict one another. Thus, the statement is most likely false.\\
    \textcolor{Plum}
{\bf It can be either because}
the context says Asians make a significant part of the poverty population, which would imply that the statement itself is false if the term significant referred to the overall number.  However, since there are more Asians than any other demographic, it could just as likely be that they are usually not poor but just make up a significant portion due to them making up a significant portion of the overall population.\\
    \bottomrule
  \end{tabulary}
  \caption{Training example prompt for the Predict-then-Explain task.
    The texts \fbox{Prediction} and \fbox{Explanations} are not part of the prompt.
    The Explain-then-Predict prompt flips the order of
\fbox{Prediction} and \fbox{Explanations}. The Predict-only prompt does not include
\fbox{Explanations}.
  }
  \label{tab:prompt}
\end{table}

We experiment with two orderings of the prediction and explanation in the
prompt: Predict-then-Explain vs.\ Explain-then-Predict (a.k.a.\ chain-of-thought
prompting). \citet{ye_unreliability_2022} found no clear winner between the two approaches for reasoning tasks like NLI.
Each prompt includes 16 training items, which is the highest number of items that
fit within the length limit when the prompt includes explanations.
We use the longest explanation for each label for training to increase lexical variability.

To investigate the effectiveness of explanations,
we also prompt the model without any explanations (Predict-only) with the same 16 training items.
However, since the models prompted with explanations receive strictly more information, we also experiment with providing additional 16 training items (Predict-only-extra-train) to compensate for the loss of information.%

We first developed the prompts on 20 items that are not used for training or testing.
For the remaining 102 items, we created three train/test splits of sizes 32/70.
We report the mean and standard deviation over the three test sets.

\paragraph{Metrics}
To measure the distance between the predicted and the ground
truth label distributions, we report KL-divergence ($[0, \infty)$), Jenson-Shannon distance (JSD) and Total Variation
Distance (TVD) (both bounded within $[0,1]$). The lower the metrics, the better the predictions.
We include a baseline of always predicting the uniform distribution with 1/3 probability for each
label (Uniform in Table~\ref{tab:scores}).
As a reference,
to see how much variation needs to be captured, we also predict the ground truth majority vote label of each item with
probability 1 (Majority Vote in Table~\ref{tab:scores}).

\paragraph{Results}
Table~\ref{tab:scores} shows the metrics for the different setups, as well as the
entropy of the predicted distributions.\footnote{98\% of the predicted
  distributions sum to 1. We normalize the distributions before computing the
  metrics.
We also smooth the predictions by adding 1e-5 to all probabilities to prevent
undefined KL on a predicted probability of 0.}
Overall, the KL and JSD results are comparable to
\citet{nie-etal-2020-learn}, who fine-tuned BERT-based \cite{devlin-etal-2019-bert} classifiers on
single-labeled data only and evaluated the softmax distributions.
Our results underperform
\citet{zhang-etal-2021-learning}, who also fine-tuned BERT-based models with
both single-labeled and soft-labeled data, suggesting room for improvement.

\begin{table}
  \resizebox{\linewidth}{!}{
\begin{tabular}{llll|l}
\toprule
{}                                  &              JSD          &              TVD         &               KL          & Entropy\\
\midrule
Explain-then-Predict                &  0.272$_{0.015}$          &  0.598$_{0.022}$         &  \textbf{0.682}$_{0.432}$ &  0.899$_{0.059}$ \\
Predict-then-Explain                &  0.278$_{0.034}$          &  0.629$_{0.096}$         &  0.982$_{0.322}$          &  0.618$_{0.066}$ \\
Predict-only                        &  0.284$_{0.016}$          &  0.626$_{0.041}$         &  1.302$_{0.518}$          &  0.579$_{0.044}$ \\
Predict-only-extra-train            &  \textbf{0.268}$_{0.025}$ &  \textbf{0.585}$_{0.051}$&  1.147$_{0.325}$          &  0.646$_{0.022}$ \\
\midrule
Uniform                             & 0.285                     & 0.665                    & 0.348                     & 1.098\\
Majority Vote                       & 0.363                     & 0.697                    & 4.066                     & 0\\
  \midrule
  \citet{nie-etal-2020-learn}       & 0.305                     &                          & 0.665                     & \\
  \citet{zhang-etal-2021-learning}  & 0.192                     &                          & 0.180                     &0.868  \\
\bottomrule
\end{tabular}
}
\caption{Metrics for each setup and mean entropy of the predicted
  distributions. Each value is the average of three random splits (standard deviation in subscript).
  Best results from \citet{nie-etal-2020-learn} and
  \citet{zhang-etal-2021-learning} on ChaosNLI given as reference -- not directly
  comparable (with us and with each other) because different test sets are used.
}
\label{tab:scores}
\end{table}

\paragraph{Effect of explanations}
Explain-then-Predict slightly outperforms the
setups with the same amount of training data. The difference in KL is larger but may not be meaningful (see below).
Doubling the amount of training items  improves some metrics: Predict-only-extra-train outperforms Explain-then-Predict on JSD and TVD.

The entropy of the prediction shows a larger difference between the models.
In particular, Explain-then-Predict predicts high entropy
distributions: predicting with lower probabilities for all labels, leading to better KL and TVD with the ground truth distributions.
Predict-only tends to be over-confident: for 40\% of the items, it incorrectly predicts some labels to have 0 probability.
This suggests that when GPT-3 generates the explanations first,
especially before the labels (as in Explain-then-Predict),
it considers more labels to be possible, leading to more uniform distributions.

\paragraph{Difference in KL is not meaningful}
The goodness on the KL-divergence
correlates with the entropy of the predicted distributions, with the Uniform
baseline scoring best.
\citet{nie-etal-2020-learn} reported a similar strong performance from the
Uniform baseline.
Instead of interpreting the pattern as suggesting that the uniform baseline is strong,
we argue that the KL-divergence definition actually explains this pattern.
KL-divergence of the ground truth $y$ from the prediction $\hat{y}$
KL$(y || \hat{y}) = H(y, \hat{y}) - H(y)$, where $H(y, \hat{y})$ is the
cross-entropy between $y$ and $\hat{y}$,
which is also the surprisal of $\hat{y}$, weighted by $y$.
When a label $x$ has nonzero ground truth probability $y(x) > 0$,
the cross-entropy $-y(x)\log\hat{y}(x)$ is smaller
when the surprisal of the prediction $-\log\hat{y}(x)$ is smaller,
which is when $\hat{y}(x)$ gets larger.
Therefore, a prediction that incorrectly predicts the label to have a 0
probability gets an exponentially larger KL than
predicting the uniform probability.
JSD, on the other hand, takes the average distribution of $y$ and $\hat{y}$ and
therefore upweights the 0-probability predictions in $\hat{y}$ and lessens the penalty.
This suggests that
the difference in KL-divergence may not be meaningful if models are performing
worse than the uniform baseline, which is the case here.

\begin{table}
  \centering
  \footnotesize
  \begin{tabulary}{1.0\linewidth}[h]{p{\linewidth}}
    \toprule
    1 \ContextStatement{
Although it is a significant part of the poverty population, Asians historically have not been able to participate in the services and programs available to the poor, he said.}{Asians are usually not poor.}\\
    \textcolor{Plum}
    {\bf It can be true because} the context states that Asians are a significant part of
    the poverty population, which implies that they are not usually poor.\\
    \textcolor{Plum}{
{\bf It can be either because}} the context does not specify whether or not Asians are usually poor or not, so it is impossible to know for sure.\\
    \textcolor{Plum}{
{\bf It can be false because}} the context states that Asians are a significant part of the poverty population, which implies that they are not usually not poor.\\
    Predicted [E,N,C]: [0.7, 0.1, 0.2] \hfill Gold: [0.0, 0.15, 0.85]\\
    KL = 1.29 \hfill JSD = 0.59 \hfill TVD = 0.7\\
    Human explanations given in Table~\ref{tab:prompt}\\
    \midrule
    2 \ContextStatement{
    What a brilliantly innocuous metaphor, devised by a master manipulator to obscure his manipulations.}{The simile was created by the manipulator.}\\
    \textcolor{Plum}
{\bf It can be true because} the context stated that the master manipulator created an innocuous metaphor, which is another name for simile, indicating that the statement is true.\\
    \textcolor{Plum}
{\bf It can be either because} the context indicates that the metaphor created was innocuous, which means it is not obviously deceptive or manipulative. Therefore, the statement could be true, meaning it was created by the manipulator, or false, meaning it was not created by the manipulator.\\
    \textcolor{Plum}
{\bf It can be false because} the context does not say for sure that the metaphor was created by the manipulator, so it could be false. The context only assumes that, or suggests it.\\
    Predicted [E,N,C]: [0.67, 0.18, 0.15] \hfill Gold: [0.3, 0.4, 0.3]\\
    KL =  0.29  \hfill JSD =  0.27 \hfill TVD =  0.37\\
    Human explanations given in Table~\ref{tab:highlight}, example \ref{ex:simile}\\
    \bottomrule
  \end{tabulary}
  \caption{
    Label prediction and generated explanations from Explain-then-Predict.
    The generated explanations can be implausible or inaccurately describe the input sentences.
  }
  \label{tab:gen_exp}
\end{table}

\paragraph{Evaluation of the generated explanations}
We perform a qualitative analysis of the generated explanations,
focusing on the output of
Explain-then-Predict on one test split containing 70 items and 218 explanations.\footnote{
  For 68 items,
  the model generated explanations for all three labels.
  For the remaining 2 items, it
  generated explanations for two labels.
  5 explanations are for labels that were predicted to have 0 probability.
  }

We found that 32 items have at least one problematic explanation: 29 explanations are highly implausible and 7 explanations inaccurately describe what is stated in the premise/hypothesis.
16 of the problematic explanations are associated to the Contradiction label, and 13 to Entailment.
The mean predicted probability of the labels with problematic explanations is 0.19, while the mean true probability of those labels is 0.10. This suggests that the model not only generates poor explanations but also predict those labels to be more likely than the ground truth.

Table~\ref{tab:gen_exp} shows examples for both kinds of errors.
Example 1 has an implausible explanation for the ``true'' (Entailment) label: being a
significant part of the poverty population cannot imply that they are not usually poor.
In Example 2, the explanation for ``false'' (Contradiction) inaccurately
describes the context: the context does say that the metaphor was created (devised) by the manipulator.
The generated explanations in both examples include template-like language, and
do not describe much reasoning over the text, unlike the human explanations.
We focus only on these two kinds of errors because they are straightforward to identify.
Evaluating model-generated explanations is a challenging and open question in
general. We leave the systematic evaluation of the generated explanations, perhaps using some of
the criteria proposed by \citet{wiegreffe-etal-2022-reframing},
for future work.

\section{Conclusion}
We introduced \livenli, containing ecologically valid explanations for different NLI
labels.
We showed that the explanations are diverse in their lexical choices and in the reasons they convey.
When prompting with GPT-3,
there is still room for improvement in both predicting label
distributions and generating explanations for different labels.
We hope \livenli can be a useful resource for studying variations in text
understanding and for building NLP models that capture variation.

\bibliography{anthology,Disagreement,explainability,custom}
\bibliographystyle{acl_natbib}
\end{document}

%% file: table_example.tex
\begin{table*}[h]
  \centering
  \footnotesize
  \begin{tabular}{p{\textwidth}}
    \toprule
    \taxex{ex:pundit}
    \nliex{Most pundits side with bushy-headed George Stephanopoulos (This Week), arguing
    that only air strikes would be politically palatable.}
    {Mr.\ Stephanopoulos has a very large pundit following due to his stance on air
    strikes only being politically palatable.}
    {[0.4, 0.3, 0.3]}\\
    \QUD{Does Stephanopoulos have a very large pundit following?}\\
    \response{ex:pundit_response_E}
    \Entailment -- This statement is most likely to be true because in the context is
    stated that ``Most pundits'' would side with Mr.\ Stephanopoulos. Most pundits
    could also mean a very large pundit following. \\
    \response{ex:pundit_both_QUD}
    \Neutral -- You cannot infer that the overall number of pundits following the individual is large just because the majority of pundits follow the individual.  He could just have 2 out of 3 total pundits following him, for instance.  Furthermore, they may be following him for reasons outside his stance on air strikes.\\
    \QUD{Do pundits follow Stephanopoulos due to his stance on air strikes?}\\
    \response{ex:pundit_response_N}
    \Neutral -- George Stephanopoulos may have a follow from pundits, but it might
    not be due to his support of drones. \\
    \response{ex:pundit_response_C}
    \Contradiction -- He might have a large pundit following, but that
    would have to be for something before the current issue of air strikes since
    one event wouldn't get people a large following overnight. \\
    \midrule
    \taxex{ex:planes}
    \nliex{They encourage the view that there's nothing--from Iraqi germ weapons programs
    to Serbian atrocities--that a few invisible planes can't fix.}
    {A few invisible planes are all it takes to fix certain issues.}
    {[0.5, 0.35, 0.15]}\\
    \QUD{Can a few invisible planes fix certain issues?} \\
    \response{ex:planes_encourage}
    \textcolor{Plum}{Neutral | Contradiction} -- The context is unclear because "They are encouraging the view" does not make the following part of the context true or false.  \\
    \QUD{Are a few invisible planes all it takes to fix certain issues?}\\
    \response{ex:planes_all}
    \Neutral -- Invisible planes may solve some problems but I'm not sure they are all it takes to fix certain issues. Could be true or false.
    \\
    \midrule
    \taxex{ex:summer}
    \nliex{\parbox[l]{\linewidth}{
    for a change i i got i get sick of winter just looking everything so dead i hate
    that}}
    {I'm so sick of summer.}
    {[0, 0.35, 0.65]} \\
    \response{ex:summer_response1}
    \Contradiction -- The context is stating how one is sick of winter, not summer,
    as the statement describes.  \\
    \response{ex:summer_response2}
    \Contradiction -- The speaker hates winter because the foliage is dead, therefore he likely loves summer when everything is alive.\\
    \response{ex:summer_response_N}
    \Neutral -- The context mentions being sick of winter while the statement mentions being sick of summer. These could both be true because the same person may still complain of summer's heat. \\
    \midrule
    \taxex{ex:wax}
    \phpair{
    The original wax models of the river gods are on display in the Civic Museum.}
    {Thousands of people come to see the wax models.}\hfill
    original MNLI: [0,1,0] \ \  \livenli [E,N,C]: [.23, .73, .04] \\
\response{}
\Neutral -- The context refers to the wax model displays in the museum. The context makes no mention of the number of visitors mentioned in the statement.\\
\response{}
\Entailment -- Museums are generally places where many people come, so if the original wax models are there it is likely thousands of people will come to see them.\\
\response{}
\textcolor{Plum}{Entailment | Contradiction} -- It's unlikely a museum could stay open for very long without thousands of visitors, so it's likely true that thousands of people come to see these wax mdoels. Unless, of course, it's a big museum with many attractions more interesting than the models, in which case the statement is likely to be false. \\
    \bottomrule
  \end{tabular}
  \caption{Examples in \livenli. \textcolor{Emerald!90!black}{\textbf{P}}: Premise. \textcolor{Emerald!90!black}{\textbf{H}}: Hypothesis.
    [E,N,C]:
    the probability distributions over the labels (E)ntailment, (N)eutral,
    and (C)ontradiction, aggregated from the multilabel annotations in \livenli.
    The explanations in the Responses (Resp.) help contextualize the label distributions.
  }
  \label{tab:examples}
\end{table*}

%% file: table_taxonomy_reannot.tex
\begin{table}
  \footnotesize
  \centering
    \begin{tabulary}{1.0\linewidth}{@{}lr@{}}
\toprule
Combination of Taxonomy Categories &  Frequency \\
\midrule
Probabilistic                &         28 \\
Lexical                      &         21 \\
\textit{QUD}                          &         15 \\
Coreference                  &          6 \\
Lexical | Probabilistic      &          5 \\
Coreference | Probabilistic  &          5 \\
Imperfection | Lexical       &          5 \\
Implicature                  &          5 \\
Imperfection | Probabilistic &          4 \\
Probabilistic | \textit{QUD}          &          4 \\
No variation              &          3 \\
Coreference | Lexical        &          3 \\
Lexical | \textit{QUD}                &          3 \\
Temporal                     &          2 \\
\midrule
Appear once: \\
\multicolumn{2}{@{}p{\linewidth}@{}}{
Implicature | Interrogative; Implicature | Lexical; Interrogative; Implicature |
\textit{QUD}; Imperfection | \textit{QUD}; Interrogative | Lexical; Coreference | Temporal;
Coreference | Imperfection; Coreference | \textit{QUD};
Implicature | Interrogative; Implicature | Lexical; Interrogative; Implicature |
Lexical | Probabilistic | \textit{QUD}; Probabilistic | Temporal; Presupposition | \textit{QUD}; \textit{QUD} | Temporal
}\\
\bottomrule
\end{tabulary}
  \caption{Frequency of the taxonomy categories in \livenli. We italicize the
    new category that emerge from \livenli.
    Label variation in annotators arises for similar reasons as what \citet{qp2} hypothesized.
  }
  \label{tab:taxonomy_reannot}
\end{table}

%% file: table_highlight.tex
\begin{table*}
  \centering
  \small
  \begin{tabular}{p{\linewidth}}
    \toprule
    \textbf{Agree on label, not on highlights} \\
    \midrule
    \taxex{ex:wreck} \textbf{P}: \colorbox{yellow!20!white}{There} \colorbox{yellow!20!white}{are} \colorbox{yellow!30!white}{other} \colorbox{yellow!40!white}{reasons} \colorbox{yellow!20!white}{that} \colorbox{yellow!80!white}{wrecks} \colorbox{yellow!80!white}{cause} \colorbox{yellow!70!white}{fan} \colorbox{yellow!70!white}{excitement--e.g.}
    \textbf{H}: \colorbox{yellow!40!white}{A} \colorbox{yellow!70!white}{cause} \colorbox{yellow!50!white}{for} \colorbox{yellow!60!white}{fan} \colorbox{yellow!60!white}{excitement} \colorbox{yellow!50!white}{can} \colorbox{yellow!50!white}{be} \colorbox{yellow!80!white}{wrecks.}
    \hfill [E,N,C]: [1, 0, 0] $\alpha$ = 0.08 \\
    \taxex{ex:tommy} \textbf{P}: \colorbox{yellow!10!white}{He} \colorbox{yellow!80!white}{leaned} \colorbox{yellow!50!white}{over} \colorbox{yellow!40!white}{Tommy,} \colorbox{yellow!30!white}{his} \colorbox{yellow!50!white}{face} \colorbox{yellow!80!white}{purple} \colorbox{yellow!30!white}{with} \colorbox{yellow!80!white}{excitement.}
    \textbf{H}: \colorbox{yellow!0!white}{He} \colorbox{yellow!70!white}{hovered} \colorbox{yellow!40!white}{over} \colorbox{yellow!30!white}{Tommy,} \colorbox{yellow!10!white}{with} \colorbox{yellow!20!white}{a} \colorbox{yellow!60!white}{deep} \colorbox{yellow!70!white}{color} \colorbox{yellow!30!white}{in} \colorbox{yellow!30!white}{his} \colorbox{yellow!30!white}{face} \colorbox{yellow!20!white}{from} \colorbox{yellow!20!white}{the} \colorbox{yellow!70!white}{thrill.} \hfill
[E,N,C]: [1, 0, 0] $\alpha$ = 0.15 \\
    \toprule
\textbf{Agree on highlights, not on labels}\\
    \midrule
    \taxex{ex:simile} \textbf{P}: What a brilliantly innocuous \colorbox{yellow!90!white}{metaphor,} \colorbox{yellow!30!white}{devised} \colorbox{yellow!10!white}{by} \colorbox{yellow!10!white}{a} \colorbox{yellow!30!white}{master} \colorbox{yellow!70!white}{manipulator} to obscure his manipulations.
    \textbf{H}: The \colorbox{yellow!90!white}{simile} \colorbox{yellow!10!white}{was} \colorbox{yellow!40!white}{created} \colorbox{yellow!20!white}{by} \colorbox{yellow!30!white}{the} \colorbox{yellow!60!white}{manipulator.}\hspace*{2.27259cm}
    [E,N,C]: [0.3, 0.4, 0.3] $\alpha$ = 0.43 \\
    \Entailment -- As all metaphors are a subset of similes, the statement must be true.\\
    \Neutral -- A metaphor is not the same as a simile, which the manipulator definitely created, but they could have created a simile too.\\
    \Contradiction -- The context describes a metaphor, while the statement mentions a simile, which is a different type of speech.\\
    \bottomrule
  \end{tabular}
  \caption{Examples where agreement on either the label or the
    highlights does not imply agreement on the other. Color transparency indicates the percentage of annotators who highlighted the word.
    The agreement on highlights do not correlate with agreement on labels.
  }
  \label{tab:highlight}
\end{table*}